%% file: main.tex
\RequirePackage{amsmath}
\documentclass{llncs}
\input{packages}

\input{macros}

\begin{document}
 \author{Piotr Czapla\inst{1} \and Jeremy Howard\inst{2} \and Marcin Kardas\inst{1}}

 \institute{
  n-waves, Wrocław\\
  \texttt{piotr.czapla@n-waves.com}, \texttt{marcin.kardas@n-waves.com}
   \and
  fast.ai\\
  University of San Francisco\\
  \texttt{j@fast.ai}
}

 \title{Universal Language Model Fine-Tuning with Subword Tokenization for Polish}

 \maketitle
 
\input{abstract}
\input{intro}

\input{related}

\input{model}
\input{experiments}

\input{final}
\bibliographystyle{splncs03}
\bibliography{bibliography}
\end{document}

%% file: packages.tex
\usepackage{lmodern}
\usepackage{amssymb}
\usepackage{mathtools}
\usepackage{graphicx}
\usepackage{booktabs}
\usepackage{hyperref}
\usepackage{tikz}
\usepackage{enumerate}
\usepackage{amsfonts}
\usepackage[T1]{fontenc}
\usepackage[utf8]{inputenc}
\usepackage{xcolor}
\usepackage{algorithm}
\usepackage[noend]{algorithmic}
\usepackage{xspace}
\usepackage{graphics}
\usepackage{pgfplots}
\usepackage{tikz}
\usetikzlibrary{arrows,arrows.meta,calc,decorations.pathmorphing,decorations.pathreplacing,backgrounds,positioning,fit,petri,chains, matrix,scopes,shapes.geometric,quotes}
\usepackage{todonotes}
\usepackage{pmboxdraw} 
\usepackage[numbers,sort]{natbib}

%% file: macros.tex
\newcommand{\PrOP}{\mathbb{P}}
\newcommand{\pr}[1]{\PrOP\left(#1\right)}
\newcommand{\EOP}{\mathbb{E}}

\newcommand{\Esub}[2]{\EOP_{#1}\left(#2\right)}

\newcommand{\defeq}{\stackrel{def}{=}}
\newcommand{\crossent}[2]{H\left(#1,#2\right)}
\newcommand{\pplOP}{\mathop{\mathrm{ppl}}}

\newcommand{\pplsub}[2]{\pplOP\nolimits_{#1}\left(#2\right)}

\newcommand{\prCond}[2]{\PrOP\left(#1\,|\,#2\right)}

\newcounter{funkcyje}
\newcounter{tempcounter}

\definecolor{dark-red}{rgb}{0.4,0.15,0.15}
\definecolor{dark-blue}{rgb}{0.15,0.15,0.4}
\definecolor{medium-blue}{rgb}{0,0,0.5}
\hypersetup{colorlinks,linkcolor={dark-red},
  citecolor={dark-blue},urlcolor={medium-blue}}

\pmboxdrawsetup{
    Block/box={\texttt{0}},
}  

%% file: abstract.tex
\begin{abstract}
Universal Language Model for Fine-tuning~\cite{ulmfit} (ULMFiT) is one of the first NLP methods for efficient inductive transfer learning. Unsupervised pretraining results in improvements on many NLP tasks for English.
In this paper, we describe a new method that uses subword tokenization to adapt ULMFiT to languages with high inflection. Our approach results in a new state-of-the-art for the Polish language, taking first place in Task 3 of PolEval’18. After further training, our final model outperformed the second best model by 35\%.
We have open-sourced our pretrained models and code.\footnote{\url{https://n-waves.com/poleval2018}, \url{http://nlp.fast.ai}}
\end{abstract}

%% file: intro.tex
\section{Introduction}
Language Modeling recently gained in importance as it is being used as a base
for transfer learning in multiple supervised tasks, obtaining impressive
improvements over state-of-the-art \cite{ulmfit,elmo,transformers}. For example
the error in text classification tasks was reduced by 18\% -- 24\% \cite{ulmfit}.
More complex tasks like commonsense reasoning and question answering were
significantly improved by applying transfer learning from a Language Model
\cite{transformers}. Use of unsupervised learning and transfer learning has
the additional benefits of greatly reduced computing time and data requirements for downstream supervised tasks. In some cases data requirements were reduced by 100 times \cite{ulmfit}.

Use of transfer learning is even more important for languages such as
Polish, where access to large supervised data sets is very limited. Most of the
language models published to date are n-gram models, that do not allow for
transfer learning and are very memory hungry.

\subsection{Our contribution}
We adapt Universal Language Model Fine-Tuning (ULMFiT)
\cite{ulmfit} to handle Polish inflection with subword tokenization using
SentencePiece \cite{sentencepiece}.  We trained multiple
models on the PolEval 2018 LM dataset. Our best model achieved a perplexity of 117.7 on
the test set, resulting in first place in the competition (second place scored a perplexity of 146.7). With further tuning after the competition of the model's hyperparameters, we lowered the perplexity to 95.0. 

We hope to see the use of FastText~\cite{fasttext} as the most common way of representing text in Polish replaced with our combination of SentencePiece and ULMFiT.

%% file: related.tex
\section{Related Work}

Language models traditionally were approximated with non-parametric models
based on counting statistics. This were recently replaced with deep neural
network for popular languages like English. However most of the literature devoted to the Polish language considers n-gram models~\cite{ngrampl1,ngrampl2,ngrampl3,ngrampl4}.
Brocki et al.~\cite{lmpl1} showed that a simple neural network ($5$ context
words with $50$ dimensional embeddings and one hidden layer) greatly outperforms
a $4$-gram solution on a Polish corpus. Regardless of performance, the n-gram
models tend to be large (several dozens gigabytes for $5$-gram~\cite{ngrampl2}),
making their use in web or mobile applications infeasible. For comparison, our
best performing model is around $150$ MB without compression. Moreover
non-parametric models do not allow for transfer learning, which is the key 
to good performance on many NLP tasks.

Natural language processing tasks show the best performance
when transfer learning is applied either from an LSTM language model
\cite{ulmfit,elmo} or from self-attention language models
\cite{transformers}.

The latter may hold the most promise as has been shown to work well on advanced NLP
tasks like question answering, however, it is hard to train and
requires extensive computing power and time~\cite{char-lm}. Therefore, we decided to first adopt an LSTM based model for Polish.

LSTMs are the most widely used RNNs. Recent state of the art performance of language models can be tracked to Merity et al.~\cite{awd-lstm}, who propose a
way to efficiently use dropout in LSTM networks as well as other regularization and
performance techniques like averaged stochastic gradient descent,
or randomized-length backpropagation through time (BPTT). This work was later
extended to transfer learning and classification by \cite{ulmfit}. Transfer learning in language modeling was shown to benefit from slanted triangular learning rates and other techniques described by \cite{cyc-lr}, originally used to quickly train computer vision models with minimal resources. 

LSTM based language models can be improved with use of adaptive methods during
inference (neural cache~\cite{neuralcache} and dynamic evaluation~\cite{dyneval}).
Both methods depend on observing sequence elements after they are predicted in
order to perform adaptation. As our Polish language model is intended for
transfer learning and not just the language modeling, we intentionally ignored
any approaches that do not benefit downstream tasks.

A few papers investigate using some more sophisticated activation functions for
the output layer (e.g., mixture of softmaxes~\cite{mos} and Hebbian
softmax~\cite{hebbian}). The use of mixture of softmaxes has been criticized for
large computing and memory requirements. Whilst Hebbian softmax is a
new work that holds a promise for a better language model for downstream tasks,
it requires significant computing power. Their models where trained for 6
days with 8 P100s, while ULMFiT can be trained in around 6 to 10 hours on one
P100.

ULMFiT's approach \cite{ulmfit} contributes a number of training tactics that allow for inexpensive training of language models. It introduced a successful approach to transfer learning and fine-tuning for NLP tasks. We selected it as our base for practical reasons such as small memory footprint, quick training time and the direct applicability to other downstream tasks like sentiment analysis.

A popular approach to transfer learning explored earlier in NLP was word
embeddings. They appear in the Polish NLP space in form of word2vec~\cite{word2vecpl1,word2vecpl2} and FastText~\cite{fasttext}.
However this approach only pretrains the first layer of a model, which greatly limits its effectiveness.

All of the word embeddings before FastText were hindered by the inflection of the
Polish language, which renders most approaches to finding embeddings for full words
incapable of learning useful features. The most successful attempt was FastText, which
uses pieces of words. 

Another approach to address inflections in Polish is to use byte pair encoding~\cite{BPE},
character level language models~\cite{elmo} or unigram subword tokenization~\cite{sentencepiece}.
We used the unigram algorithm as its representation of Polish words most closely fitted the 
training pipeline of ULMFiT, and because it has shown state of the art performance in downstream tasks such as machine translation.

%% file: model.tex
\section{Model}
\subsection{Dataset}
Our language model was trained only on PolEval 2018 LM data\footnote{\url{https://n-waves.com/poleval2018/competition} - the url to the competition will most likely change in 2019 so here is an up to date redirection.}. A summary of the datasets is presented in Table~\ref{tbl:datasets}.

The vocabulary is created from all tokens appearing at least 3 times in the training data, yielding a vocabulary of $1.38$ M tokens.

\begin{table}
    \centering
    \setlength{\tabcolsep}{8pt}
    \caption{Summary of PolEval 2018 LM datasets. The tokens denoting beginning and end of sentence are not included.}
    \label{tbl:datasets}
    \begin{tabular}{lrrr}
     \toprule
     dataset & sentences & tokens & OOV rate\\
     \midrule
     train & 23.0 M & 451.8 M & 0.73\%\\
     train (dedup.) & 21.3 M & 423.9 M & 0.78\%\\
     test & 2.6 M & 50.2 M & 0.91\%\\
     test (dedup.) & 2.4 M & 48.6 M & 0.94\%\\
     \bottomrule
    \end{tabular}
\end{table}



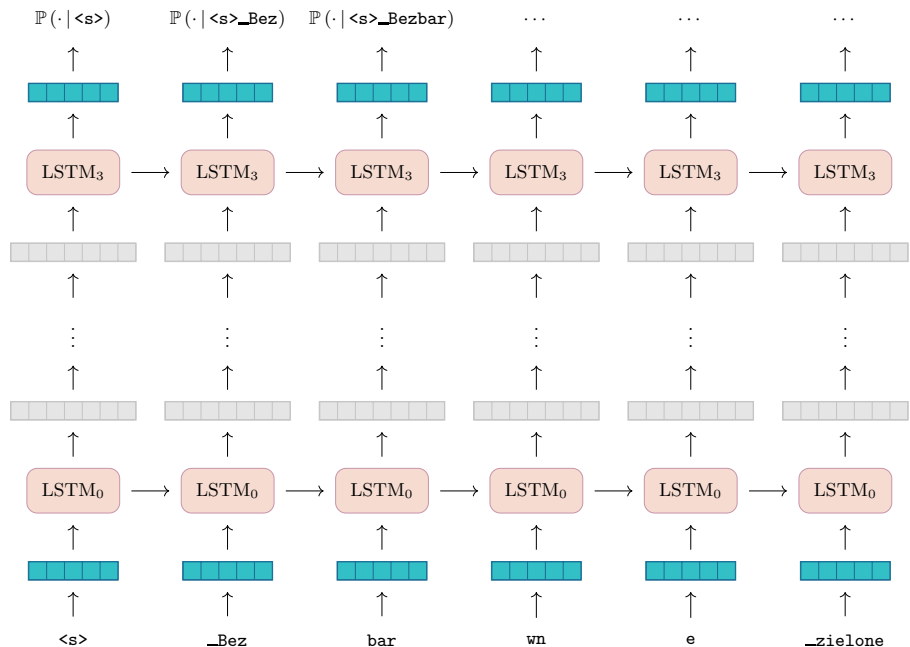
\begin{figure}[h]
    \label{fig:model}
    \input{sketch}
    \caption{An ULMFiT architecture with 4 recurrent layers.}
\end{figure}
\subsection{Subword tokenization}
Similarly as in the requirements of the competition, ULMFiT represents tokens using a fixed-size vocabulary.
Tokens not in the vocabulary are replaced by the special \texttt{<unk>} token.
However, by mapping tokens to integer identifiers we get rid of information regarding words structure. As a result, a language model operating on full words needs much more data to learn rules of highly inflected languages like Polish.

One of solutions to this problem is to use a character level model~\cite{dyneval,char-lm}.
Compared to word-based models, character level models are larger and require higher computational costs to achieve the same performance~\cite{char-vs-word}.
To gain the advantages of both approaches we trained a model working on parts of words.
The subword vocabulary is created by training a SentencePiece\footnote{\url{https://github.com/google/sentencepiece}} tokenization model.
We use a unigram segmentation algorithm~\cite{sentencepiece}.
Table~\ref{tbl:splits} shows an example of subword tokenization of a sentence for various vocabulary sizes.
An important property of SentencePiece tokenization, necessary for us to obtain a valid word-based language model, is its reversibility.
We do not use subword regularization as we decided that the available training dataset is large enough to avoid overfitting.

\begin{table}[t]
    \centering
    \setlength{\tabcolsep}{2pt}
    \caption{An example split of sentence "Bezbarwne zielone idee wściekle śpią ."
    (Colorless green ideas sleep furiously) into subword tokens using SentencePiece models differing by vocabulary sizes.
    Ratio denotes an average number of subword tokens used to encode an input token.
    The bottom part of the table was obtained by applying a lowercasing preprocessing step (see Section~\ref{sec:ulmfit}).}
    \label{tbl:splits}

    \begin{tabular}{rrl}
     \toprule
     $|V|$ & ratio & \\
     \midrule
$4k$   &   1.90 & \texttt{▁B e z bar w ne ▁zielon e ▁i de e ▁w ści ek le ▁ ś pi ą ▁.}\\
$8k$   &   1.67 & \texttt{▁Bez bar w ne ▁zielon e ▁ide e ▁w ści ek le ▁ ś pi ą ▁.}\\
$25k$  &   1.42 & \texttt{▁Bez bar wn e ▁zielone ▁ide e ▁w ście kle ▁śpi ą ▁.}\\
$50k$  &   1.34 & \texttt{▁Bez bar wne ▁zielone ▁idee ▁w ście kle ▁śpi ą ▁.}\\
$100k$ &   1.29 & \texttt{▁Bez bar wne ▁zielone ▁idee ▁w ście kle ▁śpią ▁.}\\\midrule
$4k$   &   2.04 &\texttt{<up> ▁bez bar w ne ▁ zielon e ▁i de e ▁w ści ek le ▁ ś pi ą ▁.}\\
$8k$   &   1.83 &\texttt{<up> ▁bez bar w ne ▁zielone ▁ide e ▁w ści ek le ▁ ś pi ą ▁.}\\
$25k$  &   1.61 &\texttt{<up> ▁bezbarwn e ▁zielone ▁ide e ▁w ście kle ▁śpi ą ▁.}\\
$50k$  &   1.53 &\texttt{<up> ▁bezbarwn e ▁zielone ▁idee ▁w ście kle ▁śpi ą ▁.}\\
$100k$ &   1.49 &\texttt{<up> ▁bezbarwne ▁zielone ▁idee ▁w ście kle ▁śpią ▁.}\\
     \bottomrule
    \end{tabular}
    \end{table}

We now present a formal justification of our approach. For a multiset of sentences $S=\{s_1, \ldots, s_N\}$ and LM $q\colon W^* \to [0, 1]$, the empirical perplexity per token is given by
\[
    \pplsub{S}{q} \defeq 2^{\crossent{\tilde{p}}{q}/\Esub{s\sim \tilde{p}}{|s|_W}}~,
\]
where
\[
    \crossent{\tilde{p}}{q} = -\frac1N \sum_{s\in S}\lg q(s)
\]
is an empirical cross-entropy and
\[
    \Esub{s\sim \tilde{p}}{|s|_W}=\frac1N \sum_{s\in S}|s|_W
\] is the average sentence length (in tokens).

Let $F\colon W^* \xrightarrow{1:1} V^*$ be a one-to-one mapping from sentences/sequences over tokens in $W$ into sequences over tokens in $V$. Having a LM $q_V\colon V^* \to [0, 1]$ we can create a LM $q_W\colon W^* \to [0, 1]$ with $q_W(s) = q_V(F(s))$. $F$ being injective guarantees that $Z \defeq \sum_{w\in W^*} q_W(s) \leq 1$. To make $q_W$ a valid distribution we could normalize it by $Z$ (computing of which could be infeasible) or simply assume that $q_W(\#)=1-Z$ for some additional symbol $\# \notin W$ (and $0$ for any other sequence containing $\#$). With $(F\circ q_V)(s) \defeq q_V(F(s))$ we have
\begin{align*}
    \lg \left(\pplsub{S}{q_W}\right) &= \lg\left(\pplsub{S}{F\circ q_V}\right) = \frac{\crossent{\tilde{p}}{F\circ q_V}}{\Esub{s\sim \tilde{p}}{|s|_W}}\\
    &= \frac{\crossent{\tilde{p}}{F\circ q_V}}{\Esub{s\sim \tilde{p}}{|F(s)|_V}} 
    \cdot\frac{\Esub{s\sim \tilde{p}}{|F(s)|_V}}{\Esub{s\sim \tilde{p}}{|s|_W})}\\
    &= \lg\left(\pplsub{F(S)}{q_V}\right) \cdot\frac{\Esub{s\sim \tilde{p}}{|F(s)|_V}}{\Esub{s\sim \tilde{p}}{|s|_W}}
\end{align*}
or equivalently
\begin{equation}
    \label{eqn:ppls}
    \pplsub{S}{q_W} = (\pplsub{F(S)}{q_V})^{\Esub{s\sim \tilde{p}}{|F(s)|_V}/\Esub{s\sim \tilde{p}}{|s|_W}}~.
\end{equation}

In our case, $W$ consists of $3$ control tokens (\texttt{<unk>}, \texttt{<s>} and \texttt{</s>}) and 1~378~027 tokens\footnote{Even though not all tokens in PolEval datasets are words (e.g., there are tokens consisting of punctuation marks) and some tokens produced by SentencePiece are valid words, for simplicity we call the former \textit{word tokens} and the later \textit{subword tokens}.} occurring $3$ or more times in the training data. $V$ is constructed by unigram model~\cite{sentencepiece} using SentencePiece subword tokenizer and consists of $4$ control tokens (additional \texttt{<pad>} token) and 24~996 subword tokens. For any sentence $s \in W^*$ we use the most probable tokenization as $F(s)$.
To get even better results we could sum over all possible splits of $s$. We believe, however, that the normalization factor $Z$ can be neglected as model should learn to ignore non-existent words or alternative tokenizations.

\subsection{Universal Language Model Fine-tuning}
\label{sec:ulmfit}
Our model is based on the fast.ai\footnote{\url{http://nlp.fast.ai/}} implementation of ULMFiT.
Table~\ref{tbl:models} gives details of our final submission as well as the best model trained after the competition.

\begin{table}[t]
    \centering
    \setlength{\tabcolsep}{16pt}
    \caption{Details of our submission and the best model trained after competition.}
    \label{tbl:models}

    \begin{tabular}{lcc}
     \toprule
     & PolEval submission & tuned model\\\midrule
     vocabulary size & 50K & 25 K\\
     RNN type & \multicolumn{2}{c}{LSTM}\\
     recurrent layers & \multicolumn{2}{c}{4}\\
     embeddings dimension & \multicolumn{2}{c}{400}\\
     hidden state dimension & \multicolumn{2}{c}{1150}\\
     training time & 18 epochs & 30 epochs\\
     data set used for training & $\approx$25\% & 100\%\\
     batch size & 192 & 128\\
     sampled softmax & 15 K samples & no\\
     text transforms & \multicolumn{2}{c}{none}\\
     perplexity & 117.8 & 95.0\\
     \bottomrule
    \end{tabular}
    \end{table}

\subsubsection{Data preprocessing}
Our preprocessing pipeline for the training data starts with counting occurrences of word tokens and extracting a dictionary consisting of words with at least $3$ occurrences.
The tokenized file is then deduplicated.

During development we experimented with an optional step of encoding words with an initial letter being the only capital letter. Such words are preceded with a special \texttt{<up>} token and the initial letter is lower-cased (see Table~\ref{tbl:splits} for an example). However, the experiments showed that there is no significant difference.

After the deduplication (and optional lower-casing) the full dataset is used to train a SentencePiece unigram model.
The dictionary extracted in the first step is used to remove rare (i.e., out-of-vocabulary) word tokens.
The resulting sentences are encoded by the SentencePiece model. Due to large size of the training dataset we do not use subword regularization -- each sentence is tokenized only once with the best encoding. The final dataset is randomly shuffled and split into a validation dataset (around $10$ million subword tokens) and a training dataset.

For the test dataset we optionally perform a lower-casing step, remove the out-of-vocabulary words and encode word tokens into subword tokens with SentencePiece model.
The deduplication step ensures that training and validation sets are disjoint.
However, because the test and the training datasets share some sentences (around 0.23 M / 9.29\% test sentences are present in the training dataset), the cross validation perplexity was always higher than the test one.

%% file: sketch.tex
\resizebox{\textwidth}{!}{%
\begin{tikzpicture}
  \definecolor{rnn-inner}{HTML}{f5dbd0}
  \definecolor{rnn-outer}{HTML}{c894aa}
  \definecolor{emb-inner}{HTML}{32bfc6}
  \definecolor{emb-outer}{HTML}{1b6b94}
  \definecolor{out-inner}{HTML}{e5e5e5}
  \definecolor{out-outer}{HTML}{c4c4c4}
  \tikzstyle{rnn} = [draw=rnn-outer,fill=rnn-inner,rounded corners=5,inner sep=0.25cm,outer sep=0.2cm]
  \tikzstyle{dots} = [draw=none,fill=none,outer sep=0.2cm]
  \tikzstyle{dense} = [rectangle,inner sep=0,outer sep=0.2cm]
  \tikzstyle{embs} = [fill=emb-inner,draw=emb-outer]
  \tikzstyle{rnnout} = [fill=out-inner,draw=out-outer]
  \tikzstyle{lbl} = [outer sep=0.2cm]
  \def\row#1{
    \node[rnn] (r#10) {LSTM$_#1$}; \&
    \node[rnn] (r#11) {LSTM$_#1$}; \&
    \node[rnn] (r#12) {LSTM$_#1$}; \&
    \node[rnn] (r#13) {LSTM$_#1$}; \&
    \node[rnn] (r#14) {LSTM$_#1$}; \&
    \node[rnn] (r#15) {LSTM$_#1$};
  }

  \def\mygrid#1#2{\tikz{\draw[step=0.3,#1] (0,0)  grid (0.3*#2,0.3);}}
  \def\dense#1#2#3{
    \node[dense,#2] (d#10) {\mygrid{#2}{#3}}; \&
    \node[dense,#2] (d#11) {\mygrid{#2}{#3}}; \&
    \node[dense,#2] (d#12) {\mygrid{#2}{#3}}; \&
    \node[dense,#2] (d#13) {\mygrid{#2}{#3}}; \&
    \node[dense,#2] (d#14) {\mygrid{#2}{#3}}; \&
    \node[dense,#2] (d#15) {\mygrid{#2}{#3}};
  }

  \def\dotsrow{
    \node[lbl] (s0) {$\vdots$}; \&
    \node[lbl] (s1) {$\vdots$}; \&
    \node[lbl] (s2) {$\vdots$}; \&
    \node[lbl] (s3) {$\vdots$}; \&
    \node[lbl] (s4) {$\vdots$}; \&
    \node[lbl] (s5) {$\vdots$};
  }

  \matrix[column sep={2.6cm,between origins},row sep=0.8cm,ampersand replacement=\&] (net) {
    \node[lbl] (p0) {$\prCond{\cdot}{\texttt{<s>}}$}; \&
    \node[lbl] (p1) {$\prCond{\cdot}{\texttt{<s>}\texttt{▁Bez}}$};\&
    \node[lbl] (p2) {$\prCond{\cdot}{\texttt{<s>}\texttt{▁Bezbar}}$};\&
    \node[lbl] (p3) {\vphantom{$\pr{\cdot}$}$\cdots$};\&
    \node[lbl] (p4) {\vphantom{$\pr{\cdot}$}$\cdots$};\&
    \node[lbl] (p5) {\vphantom{$\pr{\cdot}$}$\cdots$};\\

    \dense{4}{embs}{5};\\
    \row{3};\\
    \dense{3}{rnnout}{7};\\
    \dotsrow;\\
    \dense{1}{rnnout}{7};\\
    \row{0};\\
    \dense{0}{embs}{5};\\

    \node[lbl] (l0) {\texttt{<s>}}; \&
    \node[lbl] (l1) {\texttt{▁Bez}}; \&
    \node[lbl] (l2) {\texttt{bar}}; \&
    \node[lbl] (l3) {\texttt{wn}}; \&
    \node[lbl] (l4) {\texttt{e}}; \&
    \node[lbl] (l5) {\texttt{▁zielone}}; \\
  };
  \foreach \layer in {0,3}{
    \foreach \a [count=\b] in {0,...,4}{
      \draw[->] (r\layer\a) -- (r\layer\b);
    }
  }
  \foreach \x in {0,...,5}{
    \draw[->] (l\x) -- (d0\x);
    \foreach \layer/\layerp in {0/1,3/4}{
      \draw[->] (d\layer\x) -- (r\layer\x);
      \draw[->] (r\layer\x) -- (d\layerp\x);
    }
  }
  \foreach \x in {0,...,5}{
    \draw[->] (d4\x) -- (p\x);
    \draw[->] (d1\x) -- (s\x);
    \draw[->] (s\x) -- (d3\x);
  }
\end{tikzpicture}
}

%% file: experiments.tex
\section{Experiments}
We run multiple experiments on around $10$ M subword tokens of data to gain an intuition on how to tune ULMFiT hyperparameters for best performance on the Polish language. Most promising solutions were trained further on the whole training set, and the best (based on validation perplexity) was selected.
In this Section we present our findings regarding tuning various hyperparameters of the ULMFiT model.

\subsection{Results}
\input{plots}
\subsubsection{Vocabulary size}
Our experiments showed that out of all tested hyperparameters, the vocabulary size has the greatest impact on model performance on Polish language.
Unlike the English ULMFiT on full words, our vocabulary size influences how the subword tokens are formed. For a large enough vocabulary two words with the same lemma are represented as two different ids, and the similarity information is lost.
The smaller the vocabulary, the closer we get to character level models.

\subsubsection{Number of recurrent layers}
We tested our models with $3$, $4$ and $5$ recurrent layers.
Each additional layer noticeably increases memory usage of model and time necessary for a single training epoch.
On a small training dataset the $5$-layer models performed significantly worse.
We do not know whether longer training, more data or subword regularization could improve the performance relative to smaller models.
The performance of $3$-layer and $4$-layer models were almost identical on a small dataset,
but training on the full dataset proved that the latter is more capable, and achieves lower validation perplexity.

\subsubsection{Text preprocessing}

In some experiments we applied lower-casing of the initial letter of each word.
To make the transform reversible, such words were preceded by \texttt{<up>} control token (see Table~\ref{tbl:splits}).
For most of the tested vocabulary sizes and number of layers there was no noticeable difference in perplexity, with an exception of 100 K tokens, where lower-casing resulted in degraded performance.



%% file: plots.tex
\definecolor{bblue}{HTML}{4F81BD}
\definecolor{rred}{HTML}{C0504D}
\definecolor{ggreen}{HTML}{9BBB59}
\definecolor{ppurple}{HTML}{9F4C7C}

\pgfplotsset{
/pgfplots/my legend/.style={
legend image code/.code={
\draw[thick,black](-0.05cm,0cm) -- (0.3cm,0cm);%
   }
  }
}

\pgfplotsset{
    discard if not/.style 2 args={
        x filter/.code={
            \edef\tempa{\thisrow{#1}}
            \edef\tempb{#2}
            \ifx\tempa\tempb
            \else
                \def\pgfmathresult{inf}
            \fi
        }
    }
}

\pgfplotsset{compat=1.11,
    /pgfplots/ybar legend/.style={
    /pgfplots/legend image code/.code={%
       \draw[##1,/tikz/.cd,yshift=-0.25em]
        (0cm,0cm) rectangle (3pt,0.8em);},
   },
}
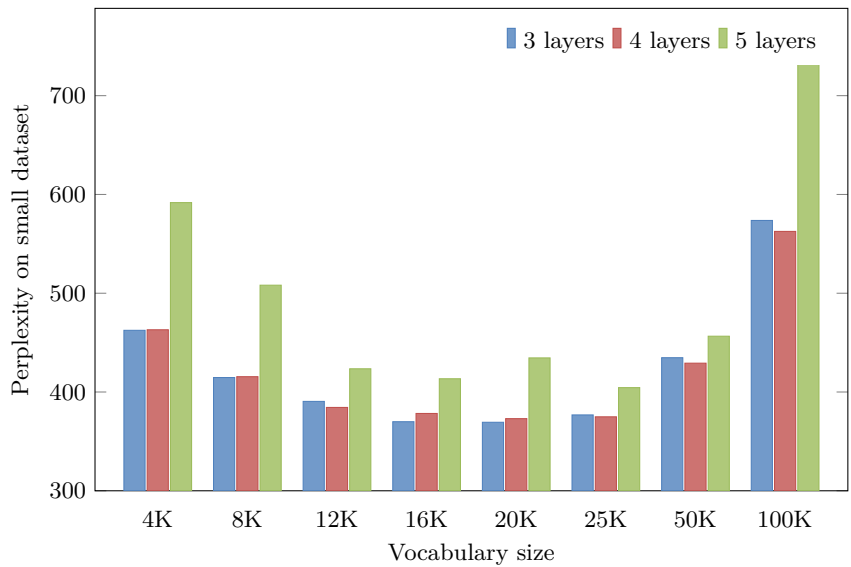
\begin{figure}[h]
    \caption{Plot showing an impact of number of recurrent layers and vocabulary size on validation perplexity. Models trained for $12$ epochs on a small dataset consisting of around $10$ M subword tokens. Models with vocabulary of size 25K and more were trained with sampled softmax (with 15K samples).}
    \label{fig:perplexity}
    \begin{tikzpicture}
        \begin{axis}[
            width  = 0.95*\textwidth,
            height = 8cm,
            major x tick style = transparent,
            ybar=2*\pgflinewidth,
            bar width=8pt,
            ylabel = {Perplexity on small dataset},
            xlabel = {Vocabulary size},
            symbolic x coords={4K,8K,12K,16K,20K,25K,50K,100K},
            xtick = data,
            scaled y ticks = false,
            ymin=300,
            legend cell align=left,
            legend columns=3,
            legend style={draw=none,nodes={inner sep=3pt}}
        ]
            \addplot[style={bblue,fill=bblue!80,mark=none},discard if not={layers}{3}] table[x=size, y=perplexity] {models.dat};
            \addplot[style={rred,fill=rred!80,mark=none},discard if not={layers}{4}] table[x=size, y=perplexity] {models.dat};
            \addplot[style={ggreen,fill=ggreen!80,mark=none},discard if not={layers}{5}] table[x=size, y=perplexity] {models.dat};
            \legend{$3$ layers,$4$ layers,$5$ layers}
        \end{axis}
    \end{tikzpicture}
\end{figure}

%% file: final.tex
\section{Final Remarks}
We showed that a subword tokenization can be used to achieve a high-performing language model for Polish, a morphologically rich language.
The presented model achieves state-of-the-art perplexity. However, we did not use the main advantage of ULMFiT, i.e., its ability for transfer learning. The natural next steps are to implement custom heads for common NLP tasks (named entity recognition, sentiment analysis) with a pretrained ULMFiT model as a backbone.